# Analysis of critical parameters of satellite stereo image for 3D reconstruction and mapping

**Rongjun Qin** [1,2], Assistant Professor
[1] Department of Civil, Environmental and Geodetic Engineering, The Ohio State University,
218B Bolz Hall, 2036 Neil Avenue, Columbus, OH 43210, USA;
[2] Department of Electrical and Computer Engineering, The Ohio State University,
205 Dreese Labs, 2015 Neil Avenue, Columbus, OH 43210, USA
qin.324@osu.edu

## ABSTRACT

Although nowadays advanced dense image matching (DIM) algorithms are able to produce LiDAR (Light Detection And Ranging) comparable dense point clouds from satellite stereo images, the accuracy and completeness of such point clouds heavily depend on the geometric parameters of the satellite stereo images. The intersection angle between two images are normally seen as the most important one in stereo data acquisition, as the state-of-the-art DIM algorithms work best on narrow baseline (smaller intersection angle) stereos (E.g. Semi-Global Matching regards 15-25 degrees as "good" intersection angle). This factor is in line with the traditional aerial photogrammetry configuration, as the intersection angle directly relates to the base-high ratio and texture distortion in the parallax direction, thus both affecting the horizontal and vertical accuracy. However, our experiments found that even with very similar (and good) intersection angles, the same DIM algorithm applied on different stereo pairs (of the same area) produced point clouds with dramatically different accuracy as compared to the ground truth LiDAR data.

This raises a very practical question that is often asked by practitioners: what factors constitute a good satellite stereo pair, such that it produces accurate and optimal results for mapping purpose? In this work, we provide a comprehensive analysis on this matter by performing stereo matching over 1,000 satellite stereo pairs with different acquisition parameters including their intersection angles, off-nadir angles, sun elevation & azimuth angles, as well as time differences, thus to offer a thorough answer to this question. This work will potentially provide a valuable reference to researchers working on multi-view satellite image reconstruction, as well as industrial practitioners minimizing costs for high-quality large-scale mapping.

## INTRODUCTION

The increasing availability and resolution of the spaceborne imagery are attracting a great attention in generating valuable 3D content of the terrain and ground object. Its advantage in highly repetitive acquisition and relatively low-cost offers a huge potential for large-scale monitoring and land-cover studies. With the development of the advanced digital dense matching algorithms, it is possible to use multi-view / stereo-view satellite images to reconstruct LiDAR (Light Detection and Ranging) comparable 3D point clouds (Gehrke et al., 2010). However, in practice it is often questioned that the results of the reconstruction are not ideal, even using the most advanced methods appearing in the lead board in benchmark testing (Scharstein and Szeliski, 2014). Often, benchmark-testing datasets were constructed under careful geometric and lighting configurations, guaranteeing minimal impact on the dense matching results (Scharstein and Szeliski, 2002). In practice, the unavoidable lighting differences and sub-optimal acquisition geometric configurations might lead to the impression of poor algorithmic performance.

In typical aerial photogrammetric acquisition missions, e.g. using Unmanned Aerial Vehicle, or aerial platforms, the acquisition can be normally controlled to capture ideal image blocks with minimal lighting differences (this is normally not a factor of concern as the acquisition are done in a relatively short period), and good geometric configuration (with designed overlaps leading to optimal base-high ratio for narrow-baseline stereo algorithms). In the case of satellite imaging, the acquisition is more restrictive in the sense of: 1) acquisition pattern; 2) acquisition time interval; 3) atmospheric effects. The flying path will need to follow the orbits, and given the fact that most of the satellite platforms carry linear array cameras (near parallel projection in the flying direction), the stereoscopic overlapping is realized through steering the camera orientations in the orbit. Normally the time interval of overlapping images on the same track (the same orbit pass) is in a matter of hours, while for overlapping images off the track (on a different orbit pass) it is in the matter of months or even years. Therefore, both the geometric and time constraints bring difficulties in digital surface model (DSM) generation: On one hand, it requires accurate steering of the camera to yield expected convergent images; on the other hand, the large time acquisition interval comes along with higher chances of physical changes on the ground, and scene being illuminated under different lighting conditions. Therefore, it is not surprising that the results of satellite stereo reconstruction are often inconsistent with the results reported in academic studies.

In understanding the performance of the stereo reconstruction, many of the existing studies only attribute the potential cause to the intersection angles (d'Angelo et al., 2014), while offering a recommended angle setup might not be necessarily sufficient for a close look at the satellite stereo 3D reconstruction. In this paper, we analyze the critical parameters/metadata associated with the geometric and radiometric/lighting of the stereo data, to demonstrate the association between these parameters and the stereo reconstruction results. We used a public dataset (will introduce in section 2) that contains 50 satellite images over the same region, with a permutation of 1200 possibilities, we are able to run stereo reconstruction for over 1,000 pairs, the metadata of which are diverse enough to draw conclusive association on critical stereo parameters, including off-nadir angles, intersection angles, sun elevation azimuth angles, etc. towards the final accuracy of the DSM.

## DATASET

The satellite images used in this work are the multi-view benchmark dataset from John's Hopkins University Applied Physics Lab's (JHUAPL) (Bosch et al., 2016; Bosch et al., 2017), containing worldview2/3 over this area across two years with a total of approximately 50 images (Figure.1 ). They are taken under various conditions containing on-track and off-track stereos with the ground resolution around 0.3-0.5 meters, with 8-band multispectral information available, a complete set of Meta information can be found on their hosting website. Each out of these 50 images is provided with their associated parameters including the acquisition off-nadir and azimuth angle, sun elevation & azimuth angle, acquisition time etc. All these datasets were performed a level-2 correction (Ortho-ready) such that there is a minimal linear camera image distortion.

Theoretically, every two of these images can form a stereo pair for stereo reconstruction. Of course, there is a possibility that some of them have very similar acquisition conditions, for example, extremely small intersection angles (e.g. < 3 degrees). In total this dataset yields 1,200 possibilities for analyzing the association between the Meta parameters of the satellite stereo pairs. Moreover, this benchmark dataset provided a means of validation: over the entire region of the test, the highly accurate airborne LiDAR data were acquired for referencing. Thus, a root-mean-square error (RMSE) can be computed based on the stereo-reconstructed DSM and LiDAR data, such that the Meta parameters of the stereo pairs can be associated with the resulting accuracy of the DSM.

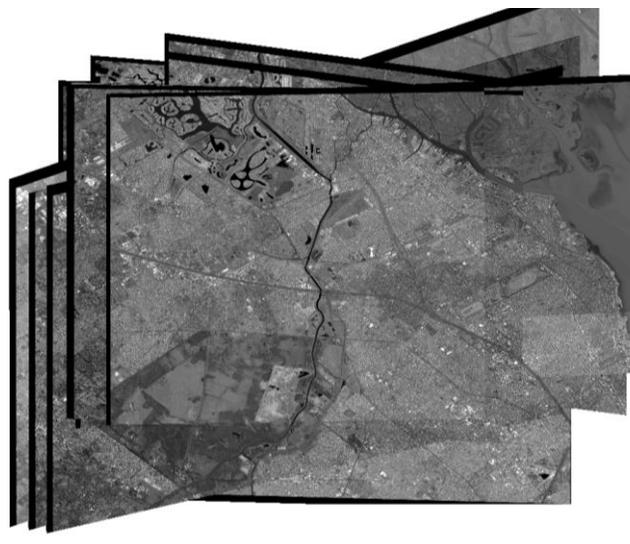

**Figure 1.** A snapshot of the 50 images. Panchromatic band, processed level-2 product. It can be seen that the angles vary, thus creating representative samples of stereo pairs

50 images yield approximately 1,200 possible stereo pairs. After eliminating those invalid stereo pairs (i.e. very small intersection angle, dramatic seasonable changes in the scene), we ended up with more than 1,000 stereo pairs.

The RSP (RPC stereo processor) (Qin, 2016) software is used to generate the DSM. RSP implements a hierarchical semi-global matching method (Hirschmüller, 2008), which presents the state-of-the-art development of dense image matching techniques. To analyze the performances of the DSM generation under different scenarios, we selected three regions of interest (1 X 1 km$^2$) are selected to perform the analysis, including scenes containing 1) moderate height commercial building region; 2) Low residential buildings and 3) high-rise commercial buildings. These regions were associated with the cropped DSM generated from LiDAR point clouds as the ground-truth for evaluation. A snapshot of these three regions is shown in Figure 2.

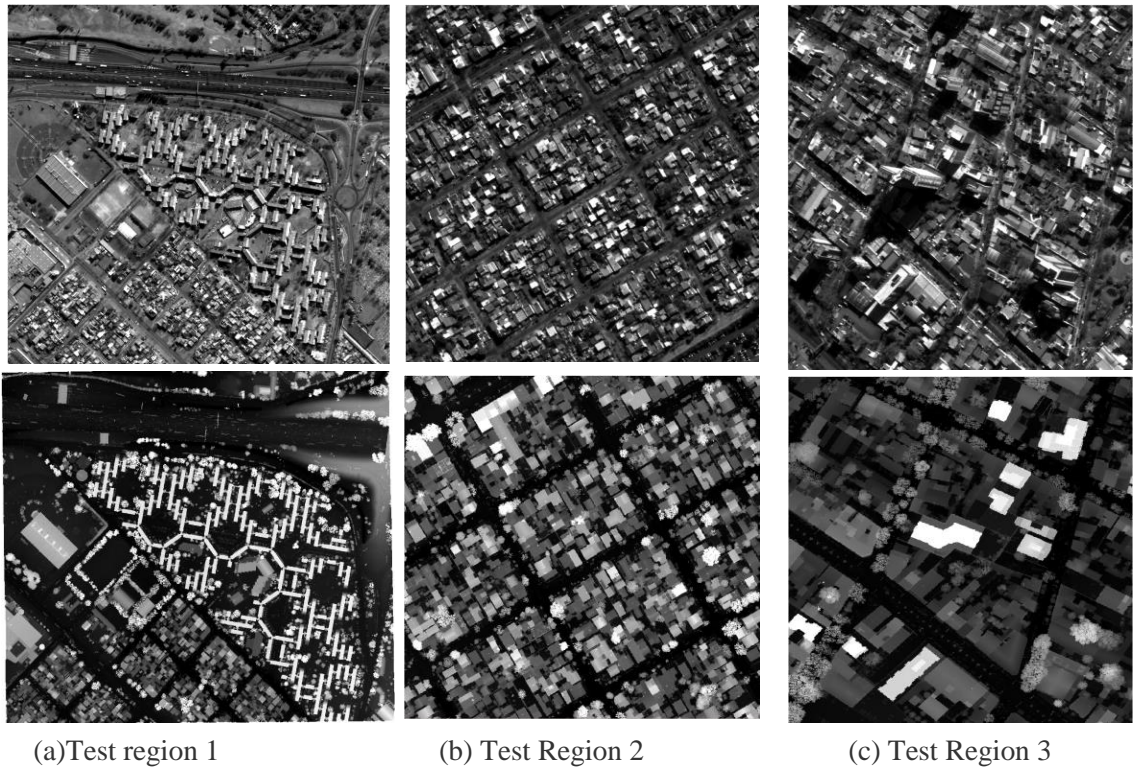

(a) Test region 1  (b) Test Region 2  (c) Test Region 3

**Figure 2**. The selected regions of interest. First row: panchromatic images, second row: DSM generated from the LiDAR point clouds. The first column (a): Test region 1, moderate height commercial building region; the second column (b): Test region 2, low residential buildings; the third column (c): Test region 3, high-rise commercial buildings.

For each stereo pair, we extract the associated metadata mainly including 1) intersection angle $\theta_{intersect}$; 2) maximal off-nadir angle of the two pairs $\theta_{maxON}$; 3) the squared difference of the sun elevation and azimuth angles $\theta_{sun-ele-azi}$. These parameters are then associated with the accuracy of the generated DSM evaluated through the corresponding LiDAR point clouds.

The relative orientation of each pair is performed through a standard RANSAC procedure using Sift/Surf features, these features are in general very robust in yielding reliable estimation of the relative orientation, where we use a simple 0th order correction factor as described in (Qin, 2016). Since there lacks an absolute orientation of the images, the alignment of the generated DSM to the LiDAR DSM were performed through an adaptive registration process considering only translation differences. Since the computed DSM is accurate in the range of 0 – 15 meters in all direction, the adaptive registration method searches the translation through a dichotomy division. To ensure the blunders not affecting the registration, we discard points that are larger than 6 meters (empirical value through the standard deviation of the DSM) in the registration processes. Two DSMs were generated for each pair, one only contains the original dense points projected onto the DSM (we call it unfilled DSM) and the other fill the occluded area with Delaunay-based triangulation (we call it filled DSM). The rationale of using both DSM for evaluation is that we want to consider the completeness. The unfilled DSM only provides the evaluation on the 3D points, which may contain very sparse and accurate points, while eventually a complete DSM without void areas are expected. Therefore, even though the interpolation of the filled DSM may bring additional modeling errors, being the wanted products it is able to leverage the completeness and accuracy.

# EXPERIMENT RESULTS AND DISCUSSION

Based on the evaluation method described in section 3, we performed our analysis on the three test areas shown in Figure 2. The association of the three parameters ($\theta_{intersect}$, $\theta_{maxON}$ and $\theta_{sun-ele-azi}$) introduced in section 3 with the accuracy of the generated DSMs is shown in Figure 3.

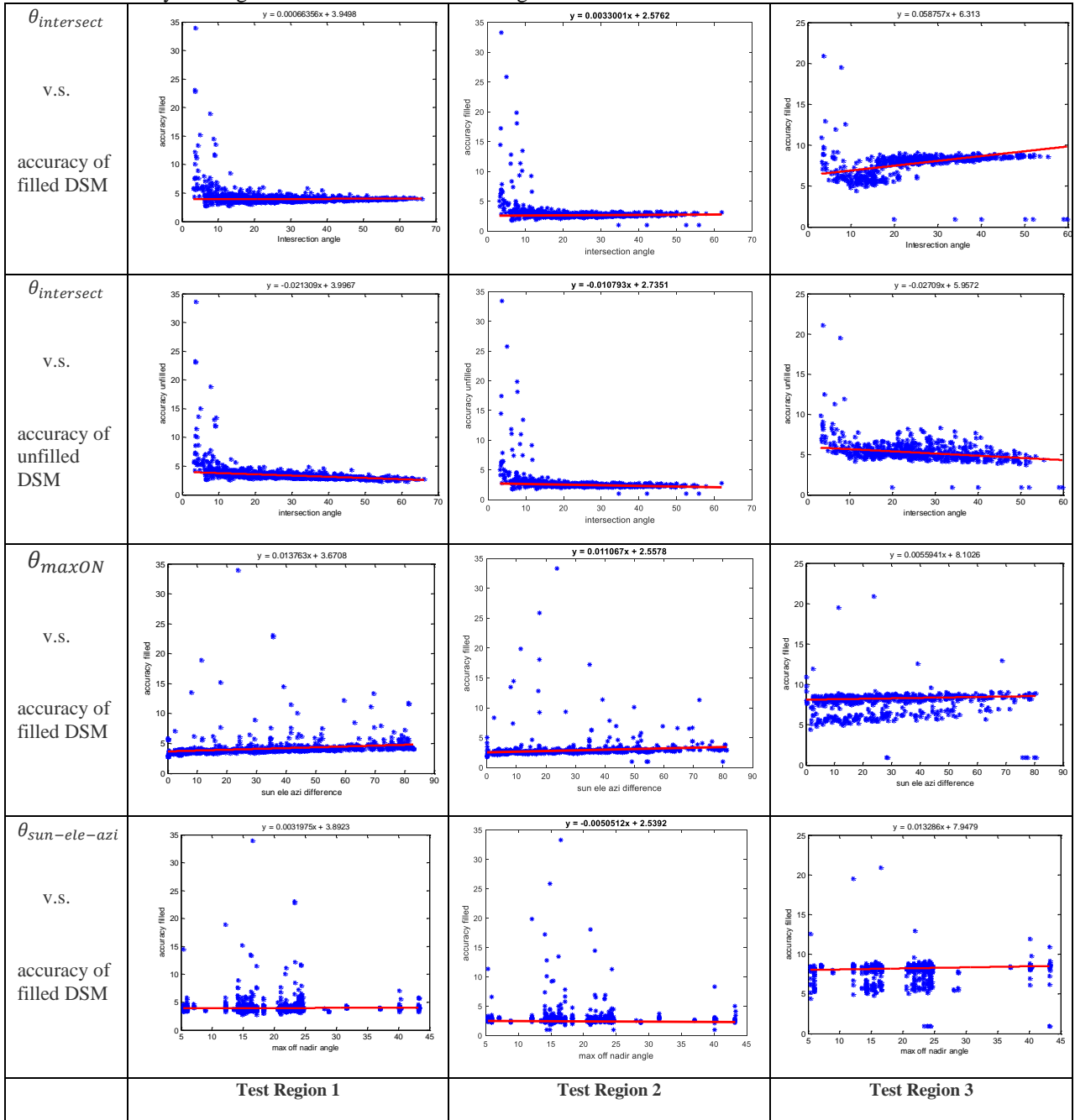

**Figure 3.** Experiment results of the Meta parameter analysis.

Each column of figure 3 shows the correlation between the metadata and the DSM accuracy. In these graphs there exist blunders, which are mainly introduced by misregistrations (due to that the DSM drifts are out of range), which can be further corrected by using more robust registration method.

Among the three test regions, the correlation between the intersection angle and the unfilled DSM clearly indicates that the larger intersection angle, the higher measurement error. This is in line with our understanding that larger intersection angles indicate larger base-high ratio resulting in better vertical accuracy given the same amount of horizontal uncertainties. However, it is expected that the larger intersection angle, the more occlusions, thus it is possible that these high accuracy points only consist of a small portion of the scene. On the other hand, the correlation between the intersection angle and the filled DSM indicates a different trend: a smaller intersection angle in general yields better results when using the filed DSM for evaluation. This indicates that, although a larger intersection angle theoretical offers higher accuracy in terms of per-point measurement, the missing parts being interpolated overall brings negative impact as the intersection angle becomes larger. This becomes particularly critical for stereo matching on scenes with high-rise buildings, evidenced by test region 3 (see Figure 3, first row, fourth column): in its scatter plot, there is an obvious discontinuity at the point of 18 degrees, from where onwards the error becomes particularly high. This is because the high-rise buildings, creating large parallax leading to large occlusions and non-matched areas, brings down the accuracy significantly, thus in such case a smaller intersection angle is of particular importance.

We also observed that the difference between the sun angles (including sun elevation and azimuth angles) when acquiring the two images, plays an important role impacting the accuracy of the generated DSM. Figure 3, the third row shows their impact to the final DSM accuracy: it clearly indicates that the larger the differences of the sun angle, the larger the error of the DSM is. This can be interpreted as: since the ground surface might not follow a Lambertian, and sometimes be a mixture of complicated specular and anisotropic surfaces, a slight lighting condition change (sun angles differ) might result in completely different textures in some region, thus leading to matching failures. Its impact, if comparing to the intersection angles (first row) is even more significant (the fitted line has steeper slope). Finally, the maximal off-nadir angle to the DSM accuracy does not show a very strong correlation: the fitted line in the experiments of the first two regions are almost flat. Although the line in the experiment of the test region 3 might seemly show a trend, its scattered points however, spread out widely in the Y-axis direction, thus leading to its prediction less reliable.

## CONCLUSION

The past study regards the intersection angle between a stereo pair plays an important role in determining the quality of the DSM generation. Many of the most recent development of dense matching algorithms favor a narrow baseline, while as a traditionally geometrical analysis, larger intersection angle (larger base-high ratio) offers higher vertical accuracy. This paper addressed this question by providing an analysis on the filled/unfilled DSM generated from stereo pairs with different intersection angles, showing that to achieve a consistent and complete digital surface model, it is recommended to have a smaller intersection angle. We also demonstrate that the differences of the sun angles at the acquisition time play an equivalent (and sometimes more important) role as the intersection angle for consideration. When selecting satellite imagery for stereo reconstruction, the sun angle differences should be accounted, and a quick guideline would be: to minimize their differences as much as possible.

## ACKNOWLEDGEMENT

The study is partially supported by the ONR grant (Award No. N000141712928). The satellite images used in this study are provided by John Hopkins University Applied Physics Lab, through the IARPA 3D Challenge.